\documentclass[lettersize,journal]{IEEEtran}
\usepackage{amsmath,amsfonts}
\usepackage{array}
\usepackage{textcomp}
\usepackage{stfloats}
\usepackage{bm}
\usepackage{url}
\usepackage{tabularx}
\usepackage{multirow}
\usepackage{adjustbox}
\usepackage{tabularx}
\usepackage{booktabs}
\usepackage{multirow}
\usepackage{amssymb}
\usepackage{verbatim}
\usepackage{graphicx}
\usepackage{cite}
\usepackage{subcaption}
\usepackage{xcolor}
\usepackage{color}
\usepackage{colortbl}
\usepackage{flushend}
\definecolor{RowColor}{rgb}{0.95, 0.95, 1}
\usepackage[ruled,vlined]{algorithm2e}
\usepackage{lineno,hyperref}
\SetKwInput{KwInput}{Input}                
\SetKwInput{KwOutput}{Output}              
\SetKwInput{KwInit}{Initialization}         

\newcommand{\bd}[1]{\textbf{#1}}

\def\eg{\emph{e.g}.}

\begin{document}

\title{MetaSeg: Content-Aware Meta-Net for Omni-Supervised Semantic Segmentation}

\author{Shenwang~Jiang$^{*}$, 
        Jianan~Li$^{*}$, 
	    Ying Wang,
	    Wenxuan Wu,
	    Jizhou Zhang,
	    Bo Huang,
	    and~Tingfa~Xu
\thanks{S. Jiang, J. Li,  Y. Wang, J. Zhang ,and T. Xu are with Beijing Institute of Technology, 100081 Beijing, China. E-mail: jiangwenj02@gmail.com, \{ lijianan, 3120215325, zhangjizhou,  ciom\_xtf1\}@bit.edu.cn. W. Wu is with Oregon State University. E-mail: wuwen@oregonstate.edu. B. Huang is with Chongqing University. E-mail:huangbo0326@cqu.edu.cn. }
\thanks{$^{*}$Shenwang Jiang and Jianan Li are co-first authors.}}

\markboth{Journal of \LaTeX\ Class Files,~Vol.~14, No.~8, August~2021}%
{Shell \MakeLowercase{\textit{et al.}}: A Sample Article Using IEEEtran.cls for IEEE Journals}

\maketitle

\begin{abstract}
Noisy labels, inevitably existing in pseudo segmentation labels generated from weak object-level annotations, severely hampers model optimization for semantic segmentation. Previous works often rely on massive hand-crafted losses and carefully-tuned hyper-parameters to resist noise, suffering poor generalization capability and high model complexity. Inspired by recent advances in meta learning, we argue that rather than struggling to tolerate noise hidden behind clean labels passively, a more feasible solution would be to find out the noisy regions actively, so as to simply ignore them during model optimization. With this in mind, this work presents a novel meta learning based semantic segmentation method, MetaSeg, that comprises a primary content-aware meta-net (CAM-Net) to sever as a noise indicator for an arbitrary segmentation model counterpart.
Specifically, CAM-Net learns to generate pixel-wise weights to suppress noisy regions with incorrect pseudo labels while highlighting clean ones by exploiting hybrid strengthened features from image content, providing straightforward and reliable guidance for optimizing the segmentation model. Moreover, to break the barrier of time-consuming training when applying meta learning to common large segmentation models, we further present a new decoupled training strategy that optimizes different model layers in a divide-and-conquer manner. Extensive experiments on object, medical, remote sensing and human segmentation shows that our method achieves superior performance, approaching that of fully supervised settings, which paves a new promising way for omni-supervised semantic segmentation.
\end{abstract}

\begin{IEEEkeywords}
Omni-supervised segmentation, meta learning, label noise
\end{IEEEkeywords}

\section{Introduction}
\label{sec:intro}

Supervised semantic segmentation methods~\cite{deeplabv3plus2018,wang2018non,zheng2020rethinking,li2021ctnet} require substantial pixel-level annotated training data, which is labor-consuming and luxurious by now.
Nevertheless, there are many data sets with massive weak annotations and few pixel-level annotations~\cite{everingham2010pascal}. 
Consequently, the issue of omni-supervised semantic segmentation, training the segmentation model by images with different level annotations~\cite{everingham2010pascal}, has attracted increasing attention recently~\cite{kulharia2020box2seg,song2019box,dai2015boxsup}.

\begin{figure}
 \centering
  \includegraphics[width=0.95\linewidth]{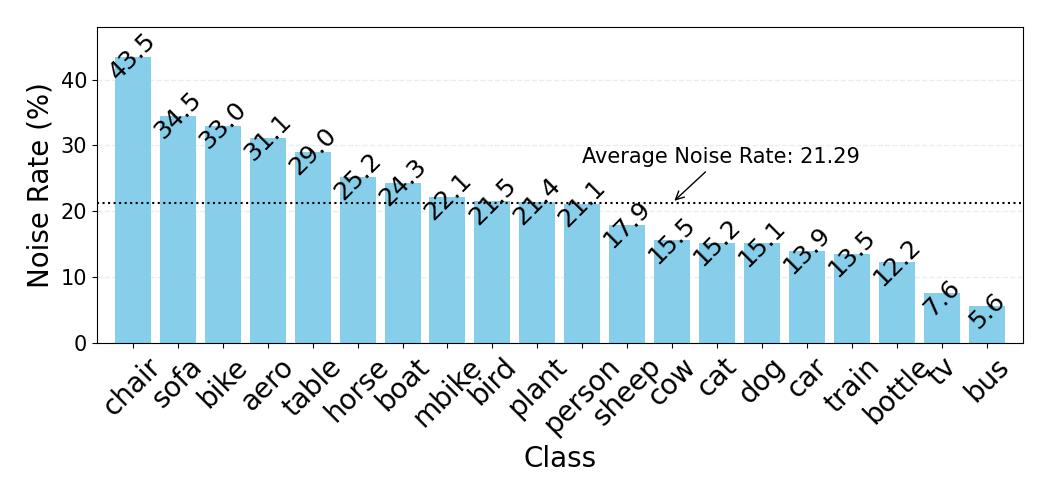}
    \caption{
    Per-class noise rate of pseudo labels generated by GrabCut on PASCAL VOC. The black dotted line represents the average noise rate on all the categories.}
    \label{fig:sta_noise_clean}
 \end{figure}

Usually, these methods  firstly transform the course label (always be bounding box) into pseudo pixel-level labels through GraphCut~\cite{rother2004grabcut} or multiscale combinatorial grouping (MCG)~\cite{arbelaez2014multiscale}, and then use the transformed labels for subsequent supervised training~\cite{li2018weakly}.
Nevertheless, the roughly generated pseudo labels are often plagued by noise whose pseudo label is inconsistent with ground truth, which largely hampers the supervised training process.
Figure \ref{fig:sta_noise_clean} shows the noise rate of pseudo labels generated by GrabCut. Clearly, noisy labels are widely existing and  unevenly distributed across all the classes, with a surprisingly large peak rate of $\bm{43.5\%}$.
Hence, the core issue of omni-supervised semantic segmentation is robust learning with noisy labels~\cite{guo2021metacorrection,kulharia2020box2seg}, which however still remains challenging and ill-solved.

Conventional approaches~\cite{kulharia2020box2seg,song2019box} mainly struggle to make the model more tolerant to label noise by introducing hand-crafted losses and carefully-tuned hyper-parameters, which suffer poor robustness to large variance and irregularity of noise.
Motivated by recent advances in meta learning~\cite{shu2019meta,wang2020training,Xu_2021_CVPR},
we are dedicated to paving a new active way for omni-supervised semantic segmentation by learning to identify noisy labels existing in the generated pseudo maps, and by weakening their impact on segmentation model training.

This then begs the question$:$ how to distinguish noisy regions from clean ones? Since convolutional neural networks (CNNs)~\cite{simonyan2014very,he2016deep} extract features layer-by-layer, high-level features often encode task-specific information yet are prone to overfitting on noisy data, and low-level features is the opposite. This inspires us that the inconsistency of low-level and high-level features is an important cue for recognizing noisy regions.

\begin{figure}
\centering
   \includegraphics[width=1.0\linewidth]{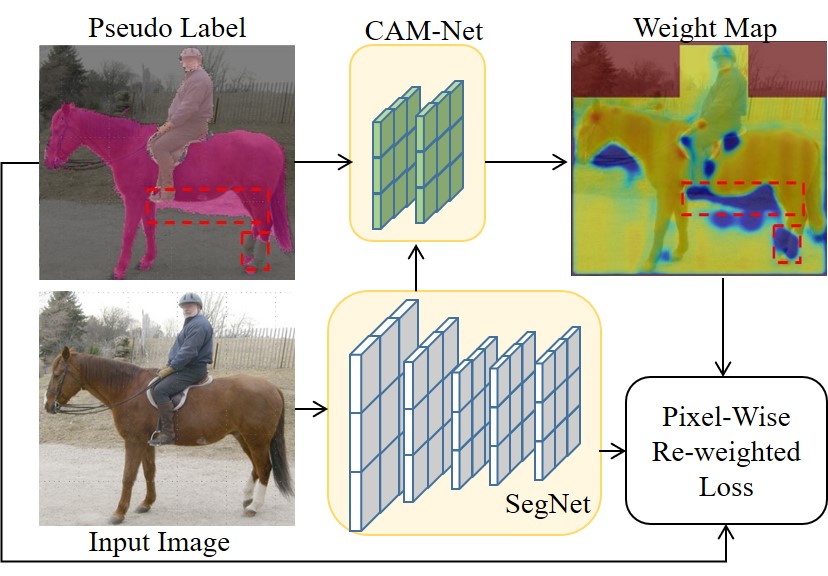}
    \caption{
    Overview of MetaSeg. The CAM-Net guides the training of SegNet by generating content-dependent pixel-wise weights for the loss function to ignore label-noisy regions (masked by red-dotted boxes).}
    \label{fig:overview}
 \end{figure}

With these in mind, this work presents a novel meta learning based framework, MetaSeg, that comprises a primary newly-designed Content-Aware Meta-Net (CAM-Net) and an arbitrary Segmentation Network counterpart (SegNet) as shown in Figure \ref{fig:overview}.
Specifically, CAM-Net first learns to recognize noisy regions by exploiting regional inconsistency among multi-level features of the input image, with the help of a series of dedicated feature strengthening operations. 
It then guides the training of SegNet by generating pixel-wise loss weights to suppress noisy regions  while highlighting clean regions existing in pseudo label maps.
Notably, CAM-Net is learnable and optimized for minimizing segmentation loss on a small-scale clean meta dataset~\cite{ren2018learning},  enabling to generate tailor-made pixel-wise weights for each input image. 

However, because segmentation models heavily rely on deep networks for accurate segmentation, there remains a major challenge, slow training speed, in applying meta learning to segmentation task.
As a remedy, inspired by the previous findings that the total meta gradient is determined mainly by only a few top layers of deep network~\cite{Xu2021FaMUS,jiang2022delving}, we introduce a novel decoupled training strategy to accelerate optimizing CAM-Net. Specifically, it freezes most bottom layers of SegNet in virtual-train~\cite{wang2020training} step and unfreezes them in actual-train~\cite{wang2020training} step.
This simple strategy significantly reduces computational overhead and thus accelerate training. 

Our MetaSeg can serve as a generic and efficient framework for omni-supervised segmentation.
Extensive experiments show that our MetaSeg achieves state-of-the-art performance for both multi- and single-class object segmentation on various datasets. 
To sum up, the main contributions of this work are:
\begin{itemize}
\item We present a novel meta learning based framework for omni-supervised semantic segmentation. 
To our best knowledge, it represents the first effort to solve omni-supervised semantic segmentation by actively identifying and weakening noisy labels through meta learning. 
\item We propose a simple yet effective decoupled training strategy to speed up the optimization of meta learning, making meta learning feasible to more applications. 
\item We establish new SOTA results for omni-supervised semantic segmentation on PASCAL VOC 2012~\cite{everingham2010pascal}, Cityscapes~\cite{cordts2016cityscapes}, ISIC 2017~\cite{Matt2017isic}, Bijie landslide dataset~\cite{ji2020landslide}, OCHuman~\cite{zhang2019pose2seg}.
\end{itemize}

\begin{figure*}[t]
\centering
   \includegraphics[width=0.8\linewidth]{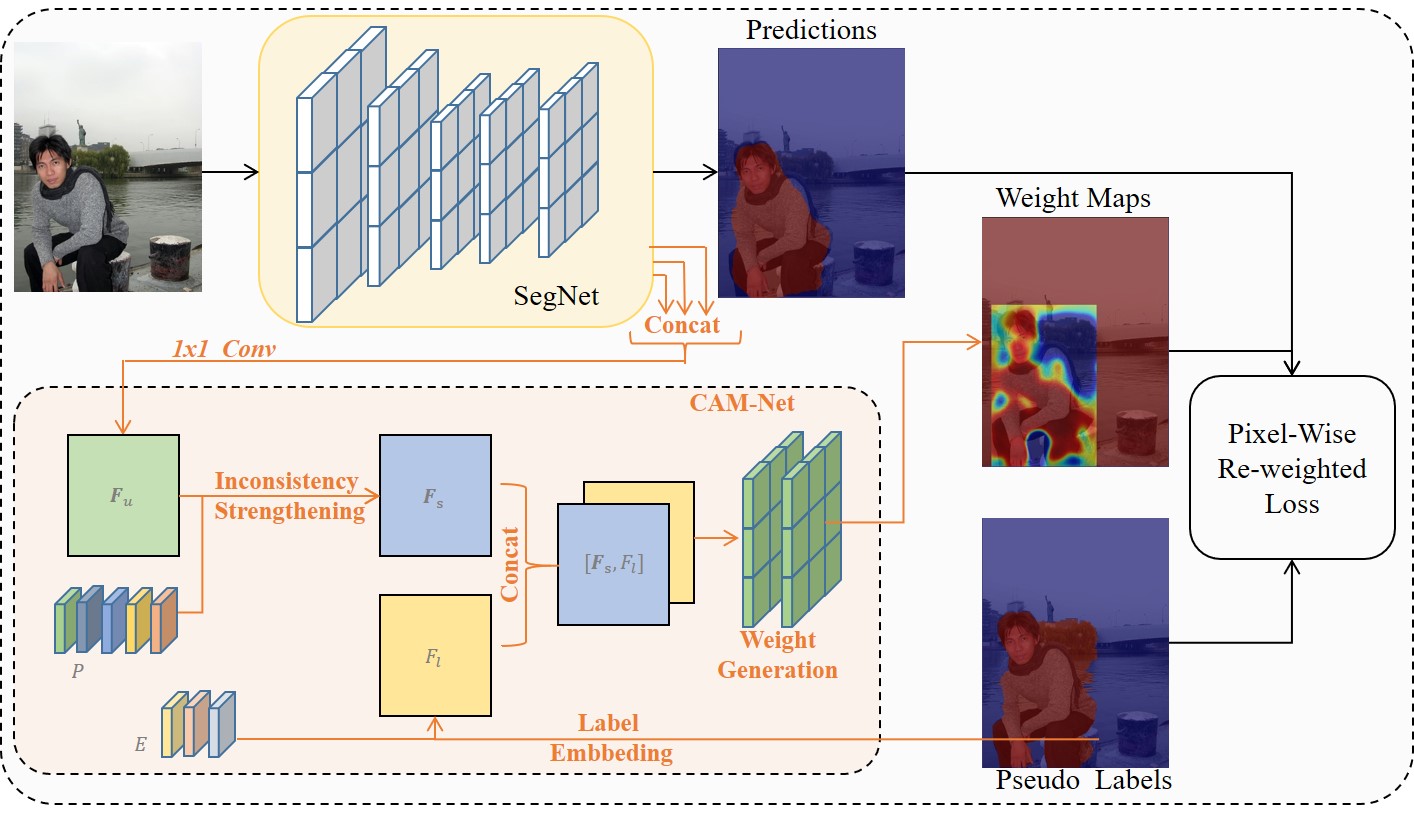}
   \caption{ Overall framework of MetaSeg. It comprises a content-aware meta-Net (CAM-Net) and a segmentation network (SegNet). CAM-Net takes both the intermediate feature of SegNet and pseudo label as input and generates content-dependent pixel-wise weight by judging the inconsistency between multi-level image features and by exploiting embedded semantic label information. The generated weights suppress noisy regions and highlight clean ones on the pixel-wise re-weighted loss for guiding the training of SegNet.}
\label{fig:framework}
\end{figure*}

\section{Related Works}
\label{sec:formatting}

Our method is mainly related with works for bounding-box-level semantic segmentation, learning with noisy labels based on meta-learning and omni-supervised semantic segmentation. 
We briefly review the relevant works as follows.

\noindent\bd{Fully-supervised Semantic Segmentation.} Semantic segmentation~\cite{deeplabv3plus2018,wang2018non,zheng2020rethinking,li2021ctnet} has made significant progress in recent years with advances in deep neural networks~\cite{he2016deep,tan2019efficientnet,dosovitskiyimage}. Notable approaches include Fully Convolutional Networks (FCN)~\cite{Zhao_2017_CVPR}, U-Net~\cite{ronneberger2015u}, and Pyramid Networks such as PSPNet~\cite{zhao2017pyramid} and DeepLab~\cite{liang2015semantic}, which use multi-scale features to capture global context. Recent works have focused on improving the efficiency of segmentation models with techniques such as attention mechanisms~\cite{li2021ctnet,9701658} and compound scaling~\cite{tan2019efficientnet}. However, fully supervised semantic segmentation still has limitations, such as the need for large amounts of accurately labeled data. As a result, there has been increasing interest in weakly- and omni-supervised~\cite{khoreva2017simple,dai2015boxsup,yi2021learning,reiss2021every} and few-shot~\cite{sunsingular,9733415,9448305,9709514} segmentation approaches, where annotations are less precise or less abundant.

\noindent\bd{Weakly-supervised Semantic Segmentation.}
The current mainstream approaches for weakly-supervised semantic segmentation, as presented in~\cite{khoreva2017simple,dai2015boxsup,song2019box,kulharia2020box2seg,oh2021background,Lee_2021_CVPR}, focus on generating pseudo labels and learning with noisy labels based on bounding-box-level annotations. SDI~\cite{khoreva2017simple} utilizes a GrabCut-like algorithm to generate pseudo labels for training a segmentation model. BoxSup~\cite{dai2015boxsup} iteratively optimizes the model between generating candidate segmentation masks and updating the segmentation model. BCM~\cite{song2019box} proposes a filling-rate-guided adaptive loss (FR-Loss) to help the segmentation model ignore noisy labels. Box2Seg~\cite{kulharia2020box2seg} predicts class-wise attention maps that guide the segmentation loss to focus on foreground regions and refine the boundaries of different classes. DFN~\cite{9705075} iteratively corrects mislabeled regions inside objects through an effective pseudo label updating mechanism and then refines high-level semantic information. BANA~\cite{oh2021background} produces pseudo labels using class activation maps~\cite{zhou2016learning} and ignores incorrect labels by a noise-aware loss (NAL). Co-T~\cite{9875225} is inspired by Co-teaching~\cite{han2018co} and trains two networks that interactively pick clean labels from one another.
Despite the significant progress achieved by the mainstream approaches for weakly-supervised semantic segmentation, they have some limitations. One of the primary drawbacks is that they rely on fixed hyper-parameters and hand-crafted losses to ignore noisy regions in pseudo labels, which may struggle with diverse noisy labels. 

\noindent\bd{Meta-learning Based Robust Learning.}
Sample re-weighting strategy is commonly adopted to alleviate the issue that deep neural networks (DNNs) are prone to overfitting to noisy labels.
L2T-DLF~\cite{wu2018learning} adopts a teacher model to generate loss functions for training the student model.
MentorNet~\cite{jiang2018mentornet} generates a suitable weight for each sample by a bidirectional LSTM network.
L2RW~\cite{ren2018learning} learns to assign weights to training samples based on their gradient directions.
Meta-Weight-Net~\cite{shu2019meta} learns an explicit weighting function for the loss of training samples.
MLC~\cite{wang2020training} directly learns a noise transition matrix T from meta-data.
FaMUS~\cite{Xu_2021_CVPR} employs an efficient layer-wise approximation method to replace the most time-consuming step in meta gradient computation.

\noindent\bd{Omni-supervised Semantic Segmentation}:
GAT~\cite{yi2021learning} proposes a graph attention network to correct noisy pseudo labels generated by class activation map with image-level annotations.
Self-correction~\cite{ibrahim2020semi} generates initial segmentation labels using instance-level annotations 
and refines the generated labels using the increasingly accurate primary model by a self-correction module during training.
Taught deep supervision~\cite{reiss2021every} integrates training information into segmentation networks and generates robust pseudo labels  for coarsely-labeled images by a mean-teacher segmentation model.
Miri~\cite{10.1007/978-3-030-33391-1_24} presents a meta-learning based method to generate pixel-wise weights for suppressing noisy regions based on loss values.
However it is limited to single-class semantic segmentation because its input is pixel-wise loss values.

As a first step, this work regards omni-supervised semantic segmentation to the problem of learning from noisy labels. We present the first generic and efficient framework for omni-supervised segmentation that ignores noisy regions while focusing on clean ones by generating content-dependent pixel-wise weights based on meta learning.

\section{Method}
Figure \ref{fig:framework} shows the overall framework of our MetaSeg. We denote the adopted semantic segmentation network as SegNet, and the weight-generation network as Content-aware Meta-Net, namely CAM-Net. 
The SegNet takes an image as input and its corresponding pseudo label as learning objective and outputs segmentation maps. 
The CAM-Net takes both the intermediate feature of SegNet and the pseudo label as input and generates content-dependent pixel-wise weights to suppress noisy regions and highlight clean ones on the input image for guiding the training of SegNet.
The SegNet is then optimized by pixel-wise re-weighting mechanism based on the generated weights.
Inspired by work for learning with noisy labels~\cite{shu2019meta,Xu2021FaMUS}, we make CAM-Net learnable by introducing meta learning to optimize it on a small amount of data with correct fine-grained annotations (meta data).

\subsection{Content-aware Re-weighting Mechanism} 
\label{sec:method:overall}
We begin by the task of supervised semantic segmentation. 
Given an input image $\bm{I} \in \mathbb{R}^{\rm H \times W \times 3}$ and its corresponding fine-grained label $\bm{Y} \in \mathbb{N} ^ {\rm H \times W \times K}$ with spatial size $\rm H \times \rm W$ and category number $\rm K$, traditional semantic segmentation method learns a model by minimizing the loss function below,
\begin{equation}
    \mathcal{L}(\bm{I},\bm{Y}|\bm{\omega}) = \frac{1}{\rm HW} \sum_i^{\rm HW} \ell(\mathcal{F}(\bm{I}_i | \bm{\omega}), \bm{Y}_i),
    \label{eq:seg_loss}
\end{equation}
where $\mathcal{F}(\bm{I}|\bm{\omega})$ denotes the segmentation model with learnable parameters $\bm{\omega}$. $\bm{I}_i$ and $\bm{Y}_i$ are the $i$-th point and its corresponding label, respectively. $\ell (\cdot)$ denotes cross-entropy loss.

However, in the case of omni-supervised semantic segmentation, because input image $\bm{I}^p$ has only coarse labels, $\bm{Y}^p$ hereby becomes its corresponding generated pseudo label in which label noise inevitably exists.
Hence, directly minimizing the loss function in Eqn. \ref{eq:seg_loss} easily causes the segmentation model to over-fit to noisy labels. 

To mitigate the impact of noisy labels, we introduce a pixel-wise re-weighting mechanism that can be used to optimize arbitrary segmentation model in a plug-and-play manner.
The core idea is to identify regions with noisy labels in the input image. For label-clean regions, we compute loss and return gradient normally; for label-noisy regions, we ignore the computed loss and corresponding gradient. The segmentation model SegNet thus 
learns mainly on the label-clean regions and does not overfit to noisy labels.

To achieve the above purpose, we need to learn a label-adaptive weight map $\bm{W}\in \mathbb{R}^{H \times W}$ that assigns a weight for each point in the image when calculating the total loss:
\begin{equation}
    \mathcal{L}_{\rm r}(\bm{I}, \bm{Y}| \bm{\omega}) =  \frac{1}{\rm HW} \sum_i^{\rm HW} \bm{W}_i \ \ell(\mathcal{F}(\bm{I}^p_i|\bm{\omega}), \bm{Y}^p_i),
\label{eq:wseg_loss}
\end{equation}
where $\bm{W}_i$ is the weight for $\bm{I}_i$. $\bm{W}_i$ has a smaller value for label-noisy regions and a larger value for label-clean regions. 

The key point is how to identify label-noisy regions in the image and learn such a weight map $\bm{W}$.  
To this end, we designed a novel Content-aware Meta-net, CAM-Net, that takes the feature map $\bm{F}$ and label $\bm{Y}^p$ of $\bm{I}^p$ as input, and outputs $\bm{W}$:
\begin{equation}
    \bm{W} = \mathcal{G}(\bm{F}, \bm{Y}^p |\bm{\theta}),
\end{equation}
where $\bm{\theta}$ is learnable parameters, which is optimized by a newly-designed decoupled alternating optimization scheme. Next, we elaborate on the detailed architecture and optimization process of CAM-Net.

\begin{figure*}[t]
\centering
   \includegraphics[width=0.95\linewidth]{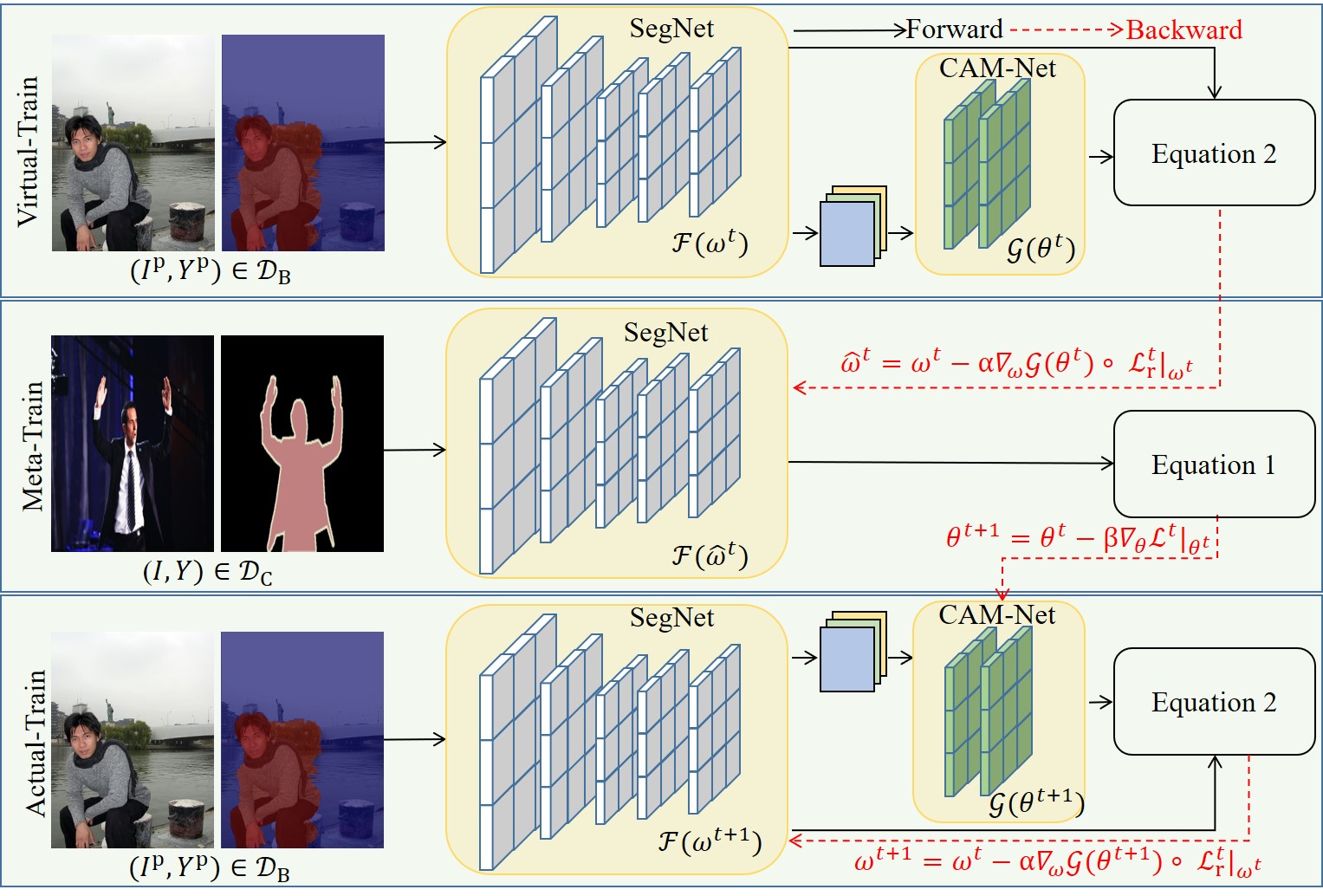}
   \caption{
   Illustration of decoupled alternating training strategy, which comprises three steps: Virtual-Train, Meta-Train, and Actual-Train.
   The SegNet is optimized on $\mathcal{D}_{\rm B}$ comprising a large number of coarsely-labeled data while the CAM-Net is trained on $\mathcal{D}_{\rm C}$ comprising a small number of finely-labeled data. $\rightarrow$ and \textcolor{red}{$\dashrightarrow$} denotes forward and backward propagation respectively.}
   \label{fig:framework_train}
\end{figure*}

\subsection{Content-Aware Meta-Net}
\label{sec:method:metanet}

As previously stated in Section \ref{sec:intro}, the inconsistency between low- and high-level features extracted by CNNs of an image can serve as an important cue to identify label-noisy regions. 
We take multi-layer feature maps of image $\bm{I}^p$ as $\bm{F}$:
\begin{equation}
  \bm{F} = \left \{ \bm{f}_i \ | \ i=3,4,5  \right \},
\end{equation}
where $\bm{f}_i \in \mathbb{R}^{\rm \frac{H}{4} \times \rm \frac{W}{4} \times \rm C}$ is the $i$-th stage resized feature map extracted by the backbone network of SegNet. 
The CAM-Net outputs label-adaptive weight map $\bm W$ by three steps: i) \textit{inconsistency strengthening}, ii) \textit{class embedding}, and iii) \textit{weight generation}.

\noindent\bd{Inconsistency strengthening}.
To strengthen the inconsistency between features from multiple layers, CAM-Net begins by fusing the information across multi-level features to obtain a fused feature map with spatial size $\rm \frac{H}{4} \times \rm \frac{W}{4}$ and channel number $\rm C$:
\begin{equation}
    \bm{F}_{u}  = \bm{\phi} ([\bm{f}_3, \bm{f}_4, \bm{f}_5]) \in \mathbb{R}^{\rm \frac{H}{4} \times \rm \frac{W}{4} \times \rm C},
    \label{eq:mf}
\end{equation}
where $[\cdot,\cdot]$ denotes concatenation operation. 
$\bm{\phi}(\cdot)$ is a linear embedding learned by $1 \times 1$ convolution.

Inspired by prototype learning~\cite{dong2018few}, we establish learnable per-category prototyping feature $\bm{P} \in \mathbb{R}^{\rm K \times R \times C}$ to store the category feature, so that each of the $\rm K$ object categories has $\rm R$ prototyping features of $\rm C$ dimension. 

Then we take the fused feature $\bm{F}_{u}$ as \textit{query} and  the prototyping feature $\bm{P}$ as \textit{key} and \textit{value} and feed them to a stack of transformer layers:
\begin{equation}
    \bm{F}_{s}  = Transformer(\bm{F}_{u}, \bm{P})\in \mathbb{R}^{\rm \frac{H}{4} \times \rm \frac{W}{4} \times \rm C}.
    \label{eq:qf}
\end{equation}
The resulting feature $\bm{F}_{s}$ enhances the inconsistency between multi-level features by comparing the $\bm{F}_{u}$ and prototyping feature  of all the categories pixel by pixel.

\noindent\bd{Label embedding.}
The pseudo label $\bm{Y}^p$ indicates which category each point belongs to and can provide useful cues for identifying noisy regions.
Since the pseudo label always be an integer on the interval $[0,K-1]$, we introduce a learnable embedding matrix $\bm E \in \mathbb{R}^{\rm K \times \rm R}$ to map $\bm{Y}^p$  from the original space to a real-valued space:
\begin{equation}
    \bm{F}_{l}  = \bm{Y}^p \bm{E} \in \mathbb{R}^{\rm \frac{H}{4} \times \rm \frac{W}{4} \times \rm R}.
    \label{eq:cf}
\end{equation}
The resulting embedded pseudo label $\bm{F}_{l}$ encodes semantic information for each point in $\bm{I}^p$, which is valuable for the identification of label noise.

\noindent\bd{Weight Generation}. To better take advantage of both cross-level feature inconsistency and point-wise semantic information for identifying noisy regions, we concatenate $\bm{F}_{s}$ and $\bm{F}_{l}$ together in the channel dimension. 
The final weight map is calculated as 
\begin{equation}
    \bm W = \bm{\sigma}([\bm{F}_s, \bm{F}_l]),
    \label{eq:if}
\end{equation}
where $\bm{\sigma}(\cdot):\mathbb{R}^{\rm \frac{H}{4} \times \rm \frac{W}{4} \times \rm 2R} \rightarrow \mathbb{R}^{\rm H \times \rm W \times \rm 1}$ is a mapping function learned by a stack of convolutions followed by upsampling and  Sigmoid activation. 
As a result, the CAM-Net generates a weight for each point in $\bm{I}^p$ based on its content, learned features and pseudo labels. 

\begin{table*}[t]
  \centering
  \renewcommand\arraystretch{1.4}
  \setlength{\tabcolsep}{1.4pt}
  \caption{Experimental settings on different datasets.}
   \label{tab:implement}%
 \begin{tabular}{c|c|c|c|c|c|c|c}
    \toprule
    \multicolumn{2}{c|}{Datasets} & PASCAL VOC 2012 & Cityscapes & ISIC 2017 & Bijie landslide & OCHuman   & Ablation Study \\
    \midrule
    \multirow{2}[1]{*}{SegNet}    
    & Frameworks & Deeplab V1~\cite{liang2015semantic} & FCN~\cite{Zhao_2017_CVPR} & FCN~\cite{Zhao_2017_CVPR} & FCN~\cite{Zhao_2017_CVPR} & Deeplabv3+~\cite{deeplabv3plus2018} & - \\
    \cline{2-8}   & Backbones & VGG-19bn~\cite{simonyan2014very} & ResNet-50~\cite{he2016deep} & U-Net~\cite{ronneberger2015u} & ResNet-50~\cite{he2016deep} & ResNet-50~\cite{he2016deep}  & - \\
    \midrule
    \multirow{2}[4]{*}{Hyper-parameters} & Iterations & 20000  & 40000  & 20000 & 20000 & 20000 & 20000 \\
\cline{2-8}          & Batch Size & 16  & 6  & 12    & 16    & 16         & 16 \\
\cline{2-8}     & Method for Pseudo Labels     & BANA~\cite{oh2021background} & Semi-supervied & External rectangles    & External rectangles    & GrabCut        & BANA \\
    \midrule
    \multirow{3}[2]{*}{Augmentations} & RandomCrop &  321$\times$321 & 769$\times$769 & False  & 128$\times$128 & 512$\times$512  & 512$\times$512 \\
\cline{2-8}          & RandomFlip & \multicolumn{6}{c}{True} \\
\cline{2-8}          & PhotoMetricDistortion & \multicolumn{6}{c}{True} \\
    \bottomrule
    \end{tabular}%
\end{table*}%

\subsection{Decoupled Alternating Optimization} 
\label{sec:method:optim}
\begin{algorithm}[t]
\caption{The Learning algorithm of MetaSeg}
\label{al:flow}
\DontPrintSemicolon
  \KwInput{Training Data $\mathcal{D}_{\rm B}$, Meta Data $\mathcal{D}_{C}$, max iterations $\rm T$.}
  \KwOutput{The parameter $\omega^T$ of Segmentation $\mathcal{F}$}
  
  \tcc{At training stage}
  \For{$t=0$ to $\rm T$}
  { 
  \tcc{Fetch mini-batch from $\mathcal{D}_{\rm B}$}
  $(\bm{I}^{\rm p},\bm{Y}^{\rm p}) \gets SampleBatch(\mathcal{D}_{\rm B})$.  \\
  \tcc{Fetch mini-batch from $\mathcal{D}_{\rm C}$}
  $(\bm{I},\bm{Y}) \gets SampleBatch(\mathcal{D}_{\rm C})$.  \\
    Virtual-Train: Update $\hat{\omega}^t$ by Equation \ref{eq:fl}. \\
    Meta-Train: Update $\theta^t$ by Equation \ref{eq:ug}. \\
    Actual-Train: Update $\omega^t$ by Equation \ref{eq:uf}.  
    }
\end{algorithm}

We next elaborate on the optimization of CAM-Net.
Let $\mathcal{D}_{\rm B}$ denote the data set comprising a large number of images $\bm{I}^{\rm p}$ with weak annotation and $\mathcal{D}_{\rm C}$ denote the meta set comprising a smaller number of images $\bm{I}$ with pixel-level annotation.
Following MLC~\cite{wang2020training}, we divide meta-learning optimization into three steps: Virtual-Train, Meta-Train, and Actual-Train, as shown in Figure \ref{fig:framework_train} and Algorithm~\ref{al:flow}.

\noindent\bd{Virtual-Train}. In the $t$-th iteration, the parameter $\bm{\omega}^t$ of the segmentation model (SegNet) is \textit{virtually} updated by loss function $\mathcal{L}_{\rm r}$  on $\mathcal{D}_{\rm B}$:
\begin{equation}
    \hat{\bm{\omega}}^{t} = \bm{\omega}^{t} - {\rm \alpha} \bigtriangledown_{\bm{\omega}}  \mathcal{L}_{\rm r}(\bm{I}^{\rm p},\bm{Y}^{\rm p}|\bm{\theta}^{t},\bm{\omega}^{t})|_{\bm{\omega}^{t}},
    \label{eq:fl}
\end{equation}
where $\rm \alpha$ is the learning rate.
The virtual-train step aims to make preparations for calculating $\mathcal{G}(\bm{\theta}^{t+1})$.

\noindent\bd{Meta-Train}. Given $\hat{\bm{\omega}}^{t}$, the parameter $\bm{\theta}$ of the CAM-Net is updated by loss function $\mathcal{L}$ on $\mathcal{D}_{\rm C}$:
\begin{equation}
    \bm{\theta}^{t+1} = \bm{\theta}^{t} - {\rm \beta} \bigtriangledown_{\bm{\theta}} \mathcal{L}(\bm{I},\bm{Y}|\hat{\bm{\omega}}^t) |_{\bm{\theta}^{t}},
    \label{eq:ug}
\end{equation}
where $\rm \beta$ is the learning rate.

\noindent\bd{Actual-Train}. After the virtual-train and the meta-train step, updated $\mathcal{G}(\bm{\theta}^{t+1})$ is obtained. Then $\bm{\omega}^t$ is \textit{actually} updated by weighted loss function $\mathcal{L}_{\rm r}$ on $\mathcal{D}_{\rm B}$: 
\begin{equation}
    \bm{\omega}^{t+1} = \bm{\omega}^{t} - {\rm \alpha} \bigtriangledown_{\bm{\omega}}   \mathcal{L}_{\rm r}(\bm{I}^{\rm p},\bm{Y}^{\rm p}|\bm{\theta}^{t+1},\bm{\omega}^{t})|_{\bm{\omega}^{t}}.
    \label{eq:uf}
\end{equation}
The segmentation model is hence optimized.

\noindent\bd{Decoupled Training}.
Nevertheless, meta learning approaches suffer super slow training.
Most of the training time is consumed for optimizing $\bm{\theta}$ as in Equation \ref{eq:ug}.
According to FaMUS \cite{Xu2021FaMUS}, $\bigtriangledown_{\bm{\theta}} \mathcal{L}(\bm{I},\bm{Y}|\hat{\bm{\omega}}^t)|_{\bm{\theta}^{t}}$ simplified as $\bigtriangledown_{\bm{\theta}} \mathcal{L}|_{\bm{\theta}^{t}}$ can be rewritten by the chain rule as, 
\begin{equation}
\begin{split}
     \bigtriangledown_{\bm{\theta}} \mathcal{L}|_{\bm{\theta}^{t}}  & = \frac{\partial \mathcal{L}}{\partial \hat{\bm{\omega}}^t}  \frac{\partial \hat{\bm{\omega}}^t}{\partial \mathcal{G}(\bm{\theta}^{t})}  \frac{\partial \mathcal{G}(\bm{\theta}^{t})}{\partial \bm{\theta}^{t}} \\
     & \varpropto \sum_i^{\rm N} \frac{\partial \mathcal{L} }{\partial \hat{\bm{\omega}}^t_i}  \frac{\partial \hat{\bm{\omega}}^t_i}{\partial \mathcal{G}(\bm{\theta}^{t})}  \frac{\partial \mathcal{G}(\bm{\theta}^{t})}{\partial \bm{\theta}^{t}},
\end{split}
    \label{eq:ug_s}
\end{equation}
where $\rm N$ represents the number of layers of SegNet.

This indicates that the amount of computation in optimizing $\bm{\theta}$ is positively related to $\rm N$.
We therefore present a new decoupled training strategy that freezes the bottom layers in virtual-train step yet unfreezes them in actual-train step.  
Equation~\ref{eq:ug_s} can be formulated as,
\begin{equation}
     \bigtriangledown_{\bm{\theta}} \mathcal{L}|_{\bm{\theta}^{t}}   \varpropto \sum_{i={\rm S}}^{\rm N} \frac{\partial \mathcal{L}}{\partial \hat{\bm{\omega}}^t_i}  \frac{\partial \hat{\bm{\omega}}^t_i}{\partial \mathcal{G}(\bm{\theta}^{t})}  \frac{\partial \mathcal{G}(\bm{\theta}^{t})}{\partial \bm{\theta}^{t}},
    \label{eq:ug_sK}
\end{equation}
where $\rm S$ is the number of decoupled layers, which determines the trade-off between speed and accuracy for meta optimization.
The decoupled training strategy can significantly reduce the time consumed in meta-train step.

\begin{figure*}[t]
\centering
   \includegraphics[width=0.9\linewidth]{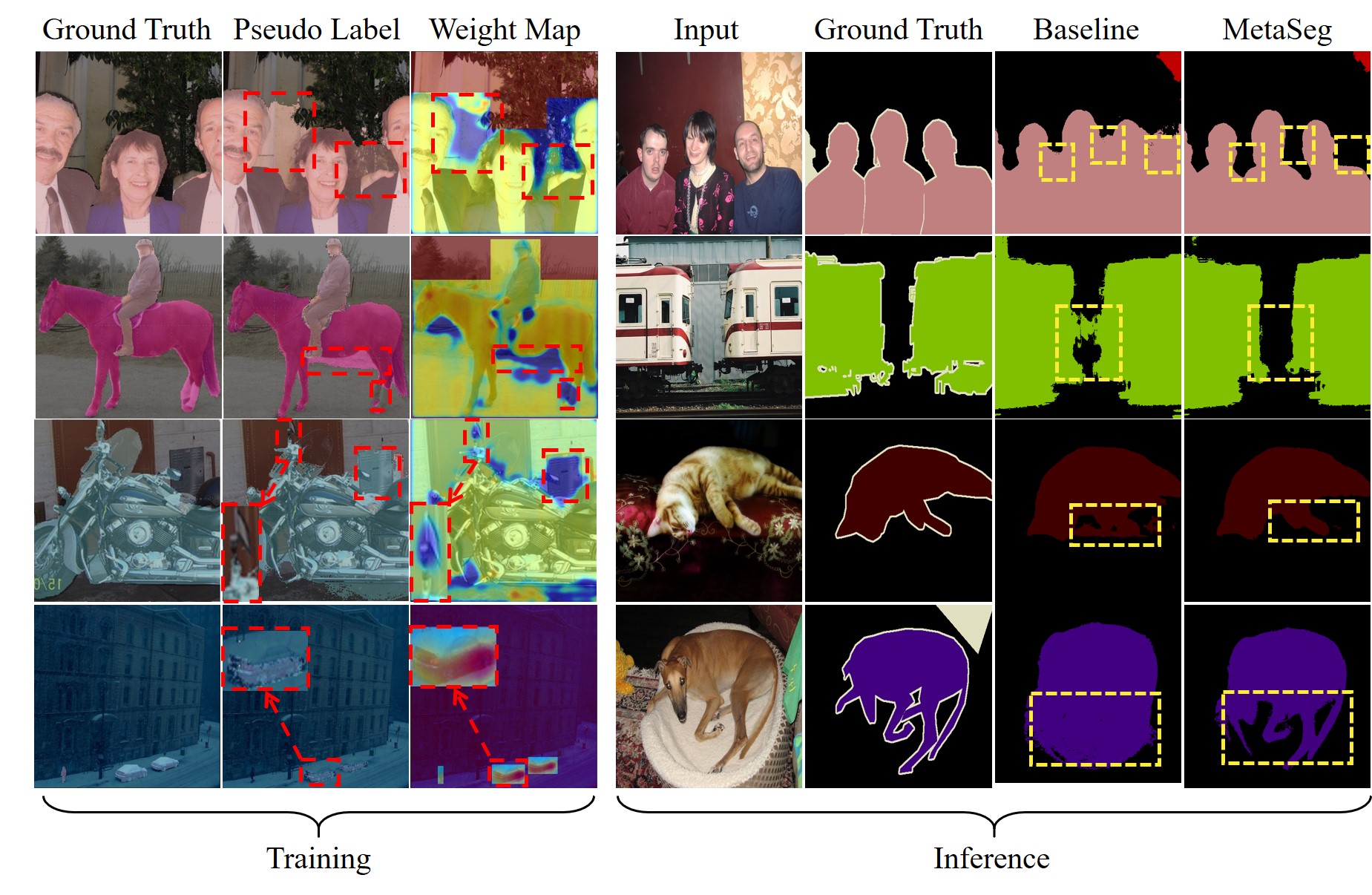}
   \caption{Visualization of ground truths, pseudo labels, and generated weights by CAM-Net at training stage (left), and qualitative results at inference stage (right) on PASCAL VOC 2012. Weight maps are overlaid on input image for clearness of spatial correspondence, with red color representing high values while blue color representing low values. \textcolor{red}{Red} dotted boxes highlight the regions with noisy labels. \textcolor{yellow}{Yellow} dotted boxes mark where our method is superior to baseline.}

\label{fig:weight}
\end{figure*}

\section{Experiments}

We evaluate our MetaSeg on a wide variety of segmentation tasks and datasets, including general object segmentation (Pascal VOC 2012~\cite{everingham2010pascal}), semantic scene segmentation (Cityscapes~\cite{cordts2016cityscapes}), medical segmentation (ISIC 2017~\cite{Matt2017isic}), remote sensing image segmentation (Bijie Landslide dataset~\cite{ji2020landslide}), and occluded human segmentation (OCHuman~\cite{zhang2019pose2seg}).
Baseline is  the SegNet trained on the images with pseudo labels in $\mathcal{D}_{\rm B}$ and then fine-tuned on the images with ground-truth masks in $\mathcal{D}_{\rm C}$.

\subsection{Implementation details.} \label{sec:exp:set}
We perform experiments on different datasets by following the experimental settings used in previous works~\cite{oh2021background,10.1007/978-3-030-33391-1_24,chen2021semisupervised}.
Specifically, for training the SegNet, we use the SGD optimizer with momentum $0.9$.
The learning rates are initialized at 0.01 and controlled by cosine annealing~\cite{loshchilov2016sgdr}.
For training CAM-Net, we use the Adam optimizer with weight decay $0$ and adopt a fixed learning rate of $0.003$.
More experimental settings on different datasets are specified in Table~\ref{tab:implement}.
Since image regions outside the bounding-box annotation are considered to be background, we manually set their corresponding weights as $1$.
All the experiments are implemented using MMSegmentation~\cite{mmseg2020} and Higher~\cite{grefenstette2019generalized}.
We use the mean Intersection-over-Union (mIoU) of all classes as evaluation metric on PASCAL VOC, Bijie Landslide, OCHuman, and Cityscapes and report the Dice score (Dice) on ISIC 2017~\cite{10.1007/978-3-030-33391-1_24}.

\subsection{General Object Segmentation} \label{sec:exp:sota_gen}

\begin{table}[t]
  \centering
  \renewcommand\arraystretch{1.4}
  \setlength{\tabcolsep}{12pt}
  \caption{Comparison with state-of-the-arts on PASCAL VOC 2012.}
  \label{tab:compare}%
    \begin{tabular}{l|l|cc}
        \toprule
        Method & Reference & Validation  & Test \\
        \midrule
        BoxSup~\cite{dai2015boxsup} & ICCV 2015 & 63.50 & 66.20 \\
        WSSL~\cite{papandreou2015weakly} & ICCV 2015 & 65.10 & 66.60 \\
        SDI~\cite{khoreva2017simple} & CVPR 2017 & 65.80 & 66.90 \\
        Souly \emph{et al.}~\cite{souly2017semi} & ICCV 2017 & 64.10 & - \\
        ADELE~\cite{liu2021adaptive} & CVPR 2022 & 65.24 & 66.25 \\
        MDC~\cite{wei2018revisiting} & CVPR 2018 & 65.70 & 67.60 \\
        Ficklenet~\cite{lee2019ficklenet} & CVPR 2019 & 65.80 & - \\
        BCM~\cite{song2019box} & CVPR 2019 & 67.50 & - \\
        \hline
        Fully Sup.~\cite{liang2015semantic} & \multicolumn{1}{c}{-} & 69.90 & 70.30 \\
         \hline
         Baseline & \multicolumn{1}{c|}{-} & 66.18 & 67.36 \\
        \rowcolor{RowColor}  MetaSeg (Ours) & \multicolumn{1}{c|}{-} & \bd{68.74} & \bd{69.17} \\
        \bottomrule
    \end{tabular}%
\end{table}%

\noindent\bd{Data and setups.}
Pascal VOC 2012~\cite{everingham2010pascal}, comprised of $20$ object categories, remains a popular dataset to evaluate SegNet in general scenarios.
The original dataset consists of 1,464, 1,449 and 1,456 images for training, validation, and testing, respectively.
Following previous methods~\cite{kulharia2020box2seg,dai2015boxsup,khoreva2017simple}, we expand the training set to $10,582$ images by adding $9,118$ extra images with official bounding-box-level annotations.
We use the $1,464$ original images to build the meta dataset $\mathcal{D}_{\rm C}$, and use the $9,118$ extra images to build
the dataset $\mathcal{D}_{\rm B}$.

\noindent\bd{Main results.}
We compare the performance of our algorithm with the SOTA omni-supervised semantic segmentation methods ~\cite{dai2015boxsup,papandreou2015weakly,khoreva2017simple,souly2017semi,song2019box,lee2019ficklenet,lai2021semi} on PASCAL VOC 2012 val and test sets in Table \ref{tab:compare}.
For fairness, the results of these methods are taken from the original papers, where the SegNet is Deeplab-V1~\cite{liang2015semantic} without any modifications and trained with same settings. 
Both in terms of val and test sets, our approach shows a large improvement over them.
Furthermore, our method yields an mIoU of 68.74\% and an mIoU of 69.17\%  on val set and test set respectively, which approaches the performance of the fully supervised model.
Compared with similar method ADELE, our method can boost the perfomance by 3.50\% mIoU, which evidences actively recognizing noisy regions and resisting them by re-weighting mechanism is more superior than resisting noise regions according to a serial of manual set hyper-parameters and by introducing a extra regularization term. 
This suggests that MetaSeg could be a promising way for omni-supervised semantic segmentation.

To qualitatively verify CAM-Net can generate tailor-made weights for each image, we visualize some examples in Figure \ref{fig:weight}.
We can clearly see that CAM-Net suppresses the noisy regions (marked with red-dotted boxes)  while highlighting the clean regions (\eg{,} person in the 1st column).
Especially, some whole tiny mislabeled parts of objects like the motorbike mirror (3rd row) can also be assigned a right weight.
In addition, consistent weights generated for objects like the sofa (2nd row) indicate  CAM-Net knows clearly the boundaries of noisy and clean regions.
We further provide qualitative comparisons in Figure \ref{fig:weight}. From it, we can see the baseline suffers from misclassification between the object and its surroundings, while MetaSeg depicts better behavior in these cases.

\begin{figure*}[t]
\centering
   \includegraphics[width=0.9\linewidth]{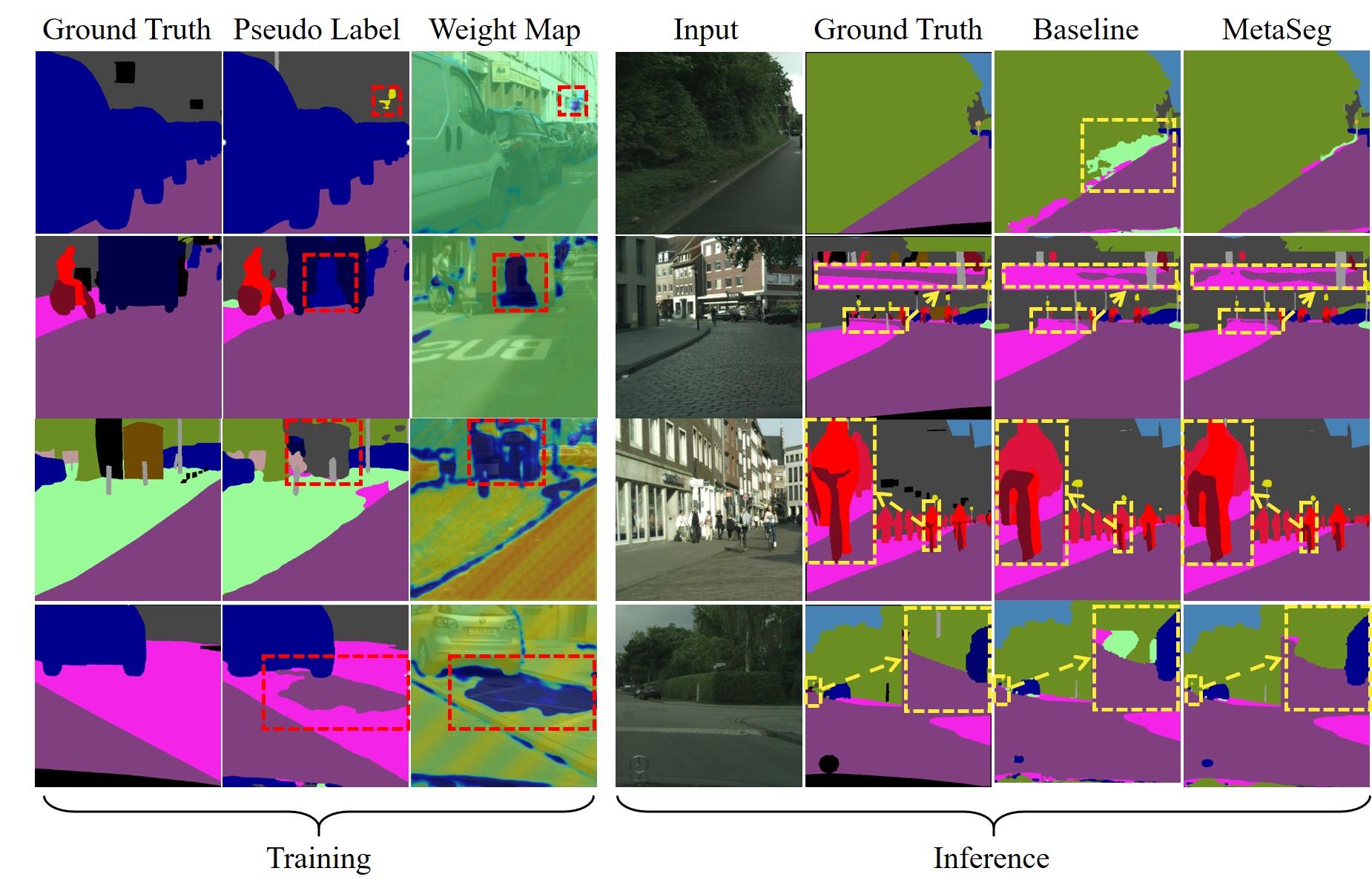}
       \caption{Visualization of ground truths, pseudo labels, and weights generated by CAM-Net at training stage (left), and qualitative results at inference stage (right) on Citysacpes. \textcolor{red}{Red} dotted boxes highlight the regions with noisy labels. \textcolor{yellow}{Yellow} dotted boxes mark where our method is superior to baseline.}
\label{fig:city}
\end{figure*}

\begin{table*}
\caption{Comparisons with state-of-the-arts on various segmentation tasks and datasets.} \label{tab:combine}
\begin{subtable}[c]{0.49\textwidth}
\centering
\renewcommand\arraystretch{1.4}
\setlength{\tabcolsep}{4.2pt}
\subcaption{Scene segmentation on Cityscapes.} \label{tab:cityscapes}
\begin{tabular}{c|cccc|c}
    \toprule
     Methods &  MT~\cite{tarvainen2017mean} & CCT~\cite{ouali2020semi} & GCT~\cite{ke2020guided} & CPS~\cite{chen2021semi} & \cellcolor{RowColor} MetaSeg  \\ 
     \midrule
     mIoU (\%) & 74.47   & 75.68  & 75.30 & 76.85 & \cellcolor{RowColor} ~\textbf{77.30}  \\
    \bottomrule
\end{tabular}%
\end{subtable}
\hfill
\begin{subtable}[c]{0.49\textwidth}
\centering
\renewcommand\arraystretch{1.4}
\setlength{\tabcolsep}{2.2pt}
\subcaption{Medical segmentation on ISIC 2017.} \label{tab:isic}
\begin{tabular}{c|ccccc|c}
    \toprule
     Methods &  Baseline &Rede$^{*}$~\cite{redekop2021uncertainty} &  Miri~\cite{10.1007/978-3-030-33391-1_24} &  Miri$^{*}$   & Fully Sup. & \cellcolor{RowColor} MetaSeg   \\
     \midrule
     Dice (\%) & 75.04 &  80.50 & 80.29 & 80.70 & 83.74  & \cellcolor{RowColor} \bd{82.76} \\
    \bottomrule
\end{tabular}%
\end{subtable}
\\
\\
\\
\begin{subtable}[c]{0.49\textwidth}
\centering
\renewcommand\arraystretch{1.4}
\setlength{\tabcolsep}{11.0pt}
\subcaption{Remote sensing segmentation on Bijie Landslide.} \label{tab:land}
\begin{tabular}{c|cc|c}
    \toprule
     Methods &  Background & Landslide & mIoU (\%)  \\
     \midrule
       Baseline     &  84.42  & 39.76 & 62.09 \\
       Fully Sup.  & 89.98 & 48.03 & 69.00 \\
       \hline
     \rowcolor{RowColor} MetaSeg (Ours)   & \textbf{87.72}     & \textbf{50.40} & \textbf{69.06} \\
    \bottomrule
\end{tabular}
\end{subtable}
\hfill
\begin{subtable}[c]{0.49\textwidth}
\centering
\renewcommand\arraystretch{1.4}
\setlength{\tabcolsep}{11.0pt}
\subcaption{Occluded human segmeantion on OCHuman.} \label{tab:ochuman}
\begin{tabular}{c|cc|c}
    \toprule
     Methods &  Background & Person  & mIoU (\%)  \\ 
     \midrule
      Baseline      & 86.53  & 65.31 &   75.92 \\
      Fully Sup.  & 86.30  & 64.86 &  75.58\\
      \hline
      \rowcolor{RowColor} MetaSeg (Ours)   &  \textbf{89.37}  & \textbf{70.19} & \textbf{79.78} \\
    \bottomrule
\end{tabular}
\end{subtable}
\end{table*}

\subsection{Semantic Scene Segmentation}  \label{sec:exp:sota_city}
\noindent\bd{Data and setups.} 
Cityscapes~\cite{cordts2016cityscapes} is a large-scale dataset for semantic scene segmentation. It consists of 2,975 training and 500 validation images each  with pixel-level annotations.
We randomly sample 744 images along with corresponding pixel-level annotations to construct meta set $\mathcal{D}_{\rm C}$.
Because some categories have no box-level annotations, we generate pseudo masks by following semi-supervised semantic segmentation settings~\cite{chen2021semisupervised,ouali2020semi,zhu2021improving}. We train the SegNet using the meta set $\mathcal{D}_{\rm C}$ and take its predictions on the train set $\mathcal{D}_{\rm B}$ as the annotations.
Finally, we evaluate our method with DeepLabv3+ on val set of cityscapes.

\noindent\bd{Main results.}
Table~\ref{tab:cityscapes} shows our CAM-Net achieves 77.30\% in the mIoU and outperforms other preeminent semi-supervised methods.
Notably, CAM-Net surpasses notable semi-supervised method CCT~\cite{ouali2020semi} by a large margin of 1.62\%, which proves the CAM-Net successfully resists the impact of noisy labels.
We visualize the weights generated by CAM-Net in right three columns sub-figures of Figure~\ref{fig:city}.
CAM-Net can accurately recognize noisy regions despite the fact that pseudo labels are generated in a semi-supervised manner. 
Under the guidance of CAM-Net, the SegNet is resistant to label noise and predicts precise segments, as shown in left four columns sub-figures of Figure~\ref{fig:city}.
These results evidence our CAM-Net can adapt to different types of annotation combinations, such as the combination of labeled and unlabeled images .

\subsection{Medical Segmentation}  \label{sec:exp:sota_med}
\noindent\bd{Data and setups.} 
Since the pixel-level semantic annotations in the task of medical image segmentation need to be labeled by scarce professional doctors, we generalize our method to medical segmentation for saving their labeling time.
We test our method for skin lesion segmentation on ISIC 2017~\cite{Matt2017isic} that comprises 2,000 training and 150 validation images.
Following Miri \emph{et al.}~\cite{10.1007/978-3-030-33391-1_24}, we randomly select 100 training images as meta set $\mathcal{D}_{\rm C}$, and the remaining training images as $\mathcal{D}_{\rm B}$.
For dataset  $\mathcal{D}_{\rm B}$, we replace the original pixel-level mask of each image with its corresponding external rectangle to mimic label noise.
We build our MetaSeg on top of classical U-Net~\cite{ronneberger2015u} and evaluate it on the valuation set.

\begin{figure*}[t]
\centering
   \includegraphics[width=0.85\linewidth]{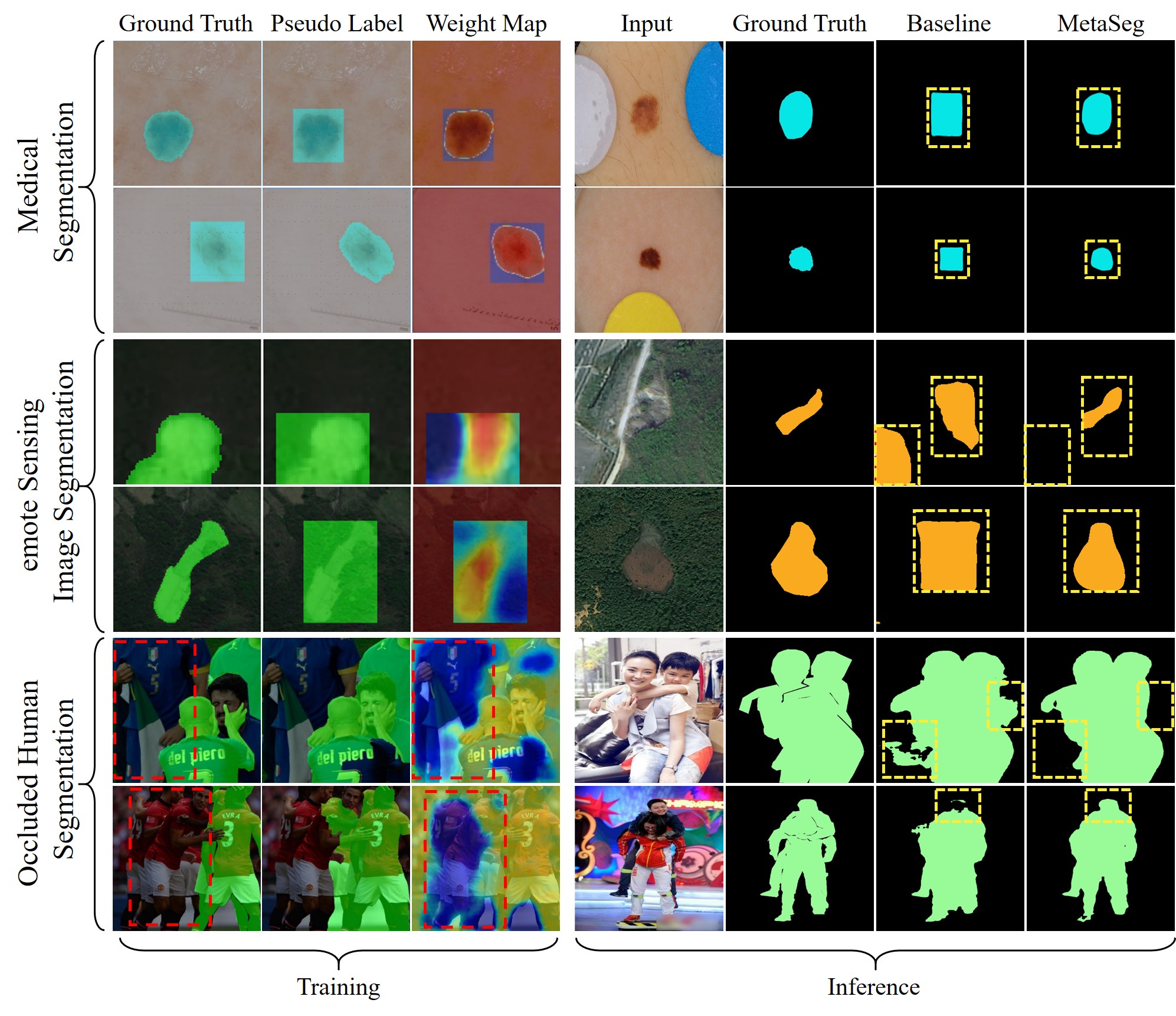}
   \caption{
   Visualization of ground truths, pseudo labels, and weights generated by CAM-Net at training stage (left), and qualitative results at inference stage (right) for medical segmentation (ISIC 2017), remote sensing image segmentation (Bijie Landslide), and occluded human segmentation (OCHuman). \textcolor{red}{Red} dotted boxes highlight regions with missing labels.  \textcolor{yellow}{Yellow} dotted boxes mark where our method is superior to baseline.}
\label{fig:combin}
\end{figure*}

\noindent\bd{Main results.}
Table \ref{tab:isic} shows the results on ISIC.
Our method achieves $82.76\%$ in the Dice score, significantly outperforming other meta learning priors~\cite{10.1007/978-3-030-33391-1_24}, even though they are optimized using precise pixel-level mask annotations. 
Notably, the performance of our method even approaches to that achieved by fully-supervised training. 
We also visualize some generated weights by our CAM-Net in 1-2 rows of Figure~\ref{fig:combin}.
One can see that our CAM-Net learns to suppress noisy regions by generating low weights on them and highlight regions with clean labels. 
These results evidence that our method can work well for medical segmentation, proving its good university.

\subsection{Remote Sensing Image Segmentation}  \label{sec:exp:sota_sensing}
\noindent\bd{Data and setups.} 
Remote sensing image segmentation is important for water quality analysis~\cite{ritchie2003remote}, mineral exploration~\cite{sabins1999remote} and disaster management~\cite{van2000remote}, however there are great lack of professionally pixel-level labeled data~\cite{wang2021dmml}.
We further evaluate our method for landslide segmentation in remote sensing imagery on Bijie Landslide dataset~\cite{ji2020landslide} that comprises 770 landslide and 2,003 non-landslide images.
We divide all the images into training, meta, and test sets in a ratio of 3:1:1. 
The noisy label maps for the images of the training set are generated by using the external rectangle of the original pixel-level mask of each image.

\noindent\bd{Main results.}
Table \ref{tab:land} shows the results on Bijie Landslide. 
Our method achieves $69.06\%$ in the mIoU, which significantly outperforms the baseline model trained with cross-entropy loss using the merged data of $\mathcal{D}_{\rm B}$ and $\mathcal{D}_{\rm C}$. 
It even performs better than the model trained in fully-supervised settings.
We suspect this is due to inaccurate pixel-level annotations in this dataset.

Similar conclusions can be drawn from the quantitative results in 3-4 rows of Figure~\ref{fig:combin}. 
When trained with the generated rectangular label masks, the baseline model is prone to predicting rectangle-shaped segmentation maps while our MetaSeg generates segmentation maps more close to the ground-truths. 
This clearly indicates that the baseline model is label-noise-agnostic and simply learns to produce results close to the given masks. In contrast, our MetaSeg can successfully distinguish foreground objects from the background and thus become more resistant to label noise.

\subsection{Occluded Human Segmentation}  \label{sec:exp:sota_human}
\noindent\bd{Data and setups.} 
We further evaluate our method for segmenting heavily occluded human on OCHuman~\cite{zhang2019pose2seg}. 
We perform this experiment because there exist a large number of human instances whose labels are missing due to heavily occlusion. These miss-labeled human instances naturally introduce label noise, making this task and dataset well suitable to evaluate our approach. 

The OCHuman dataset comprises 4,713 training, 2,500 validation, and 2,231 test images, each annotated with both bounding box and pixel-level mask for each human instance. 
We use the training images along with their bounding-box annotations as the training set $\mathcal{D}_{\rm B}$ and the validation images with their pixel-level mask annotations as the meta set $\mathcal{D}_{\rm C}$. 
We build our model on top of FCN~\cite{Zhao_2017_CVPR}.

\noindent\bd{Main results.}
As listed in Table~\ref{tab:ochuman}, our method outperforms the baseline model. It even performs better than the model trained with fully-supervised setting. One possible explanation is that the miss-labeled human instances existing in the training data introduce much label noise and thus severely affect fully-supervised training. 
Our method demonstrates high robustness against such situation by learning to assign low weights to miss-labeled instances during training, as shown in 5-6 rows of Figure~\ref{fig:combin}.

\begin{figure*}
 \centering
 \begin{subfigure}[b]{0.32\linewidth}
    \includegraphics[width=\linewidth]{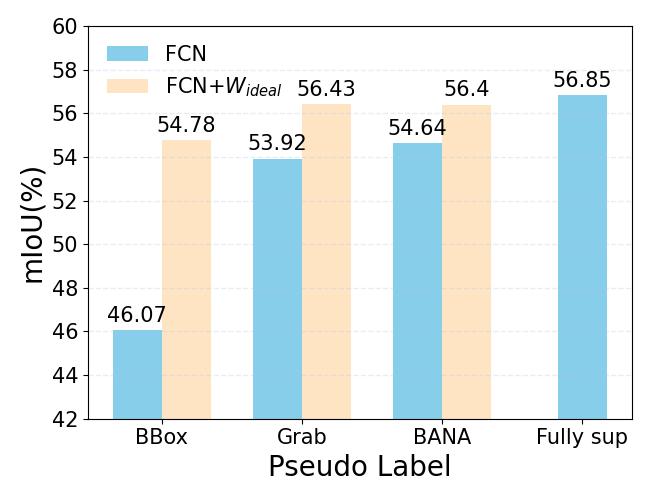}
    \vspace{-0.6cm}
     \caption{Upper-bound performance analysis.}\label{fig:pseudo}
   \end{subfigure} \hfill
   \begin{subfigure}[b]{0.32\linewidth }
   \includegraphics[width=\linewidth ]{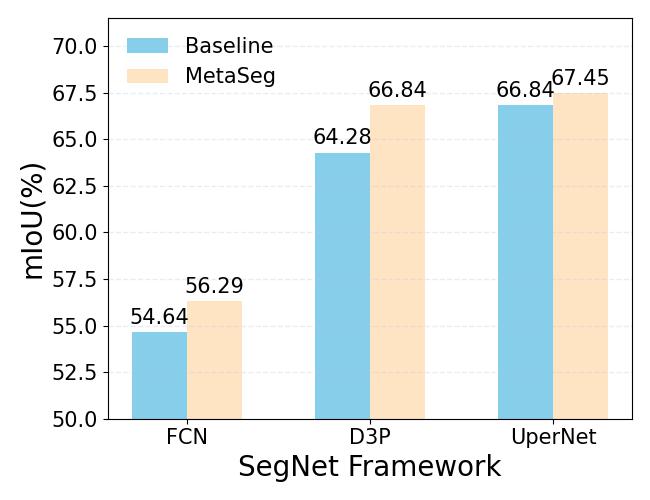}
   \vspace{-0.6cm}
   \caption{Ablation on different frameworks.}\label{fig:method}
   \end{subfigure} \hfill
   \begin{subfigure}[b]{0.32\linewidth }
   \includegraphics[width=\linewidth ]{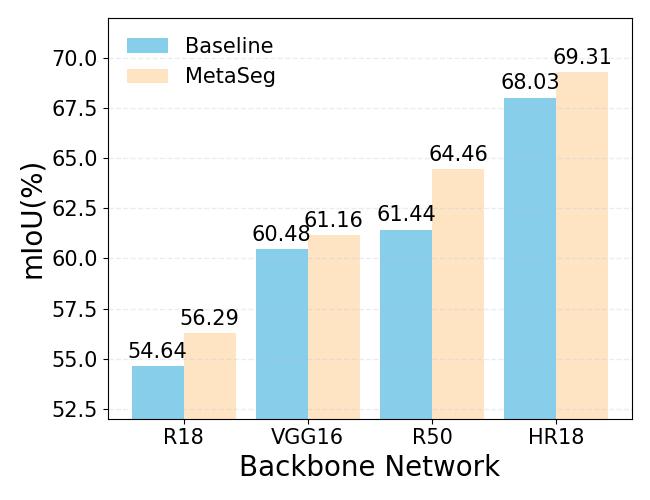}
   \vspace{-0.6cm}
   \caption{Ablation on different backbones.}
   \label{fig:backbone}
   \end{subfigure}  \\
   \begin{subfigure}[b]{0.32\linewidth }
   \includegraphics[width=\linewidth ]{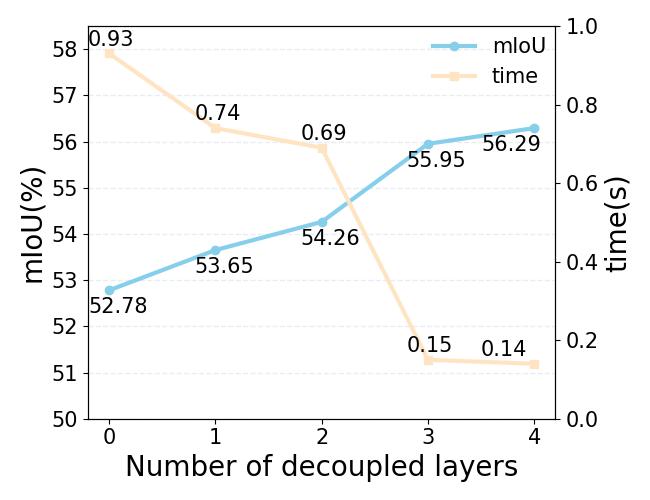}
   \vspace{-0.6cm}
   \caption{Effect of decoupled training strategy.}\label{fig:decouple}
   \end{subfigure} \hfill
   \begin{subfigure}[b]{0.32\linewidth }
   \includegraphics[width=\linewidth ]{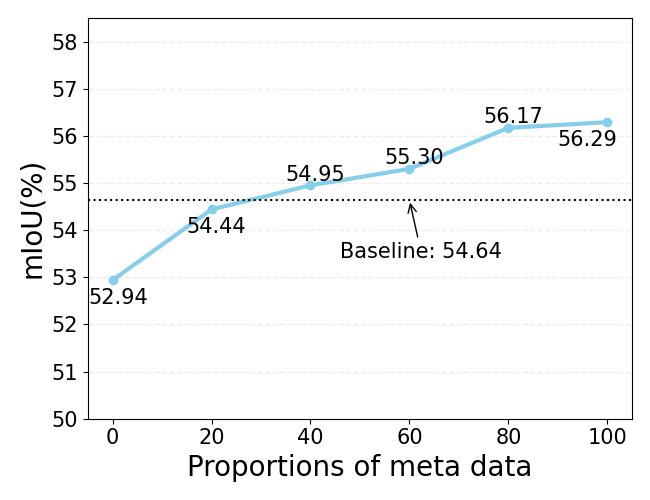}
   \vspace{-0.6cm}
   \caption{Impact of the scale of meta set.}
   \label{fig:proportions}
   \end{subfigure} \hfill
  \begin{subfigure}[b]{0.32\linewidth }
    \includegraphics[width=\linewidth ]{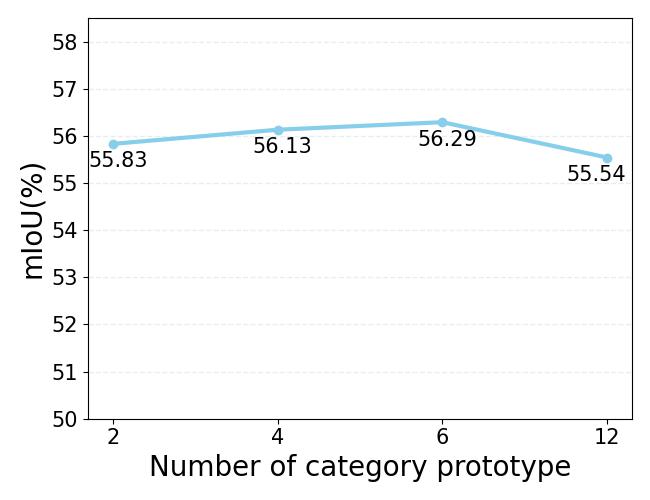}
    \vspace{-0.6cm}
     \caption{Ablation on category prototypes.}\label{fig:attr}
   \end{subfigure}
   
    \caption{Ablation studies on PASCAL VOC 2012 \textit{val} set. 
    (a): $W_{ideal}$: manually ignoring noisy regions according to ground truth.
   ``Fully sup.'': the model trained with pixel-level mask labels.}
    \label{fig:ablations}
   \end{figure*}

\subsection{Ablation Studies.} \label{sec:exp:abl} 
We perform extensive ablation studies on Pascal VOC 2012 to validate our design choices and parameter settings. 

\noindent\bd{Upper-bound analysis.}  
First, we train a SegNet based on FCN~\cite{Zhao_2017_CVPR} with ResNet18 backbone~\cite{he2016deep} using the ground-truth pixel-level mask in a fully-supervised manner. It achieves $56.85$ in the mIoU, which can be regarded as the upper limit of mIoU score of the SegNet in our experiment. 
Then we train three omni-supervised models by using the pseudo labels generated by bounding-box annotation, GrabCut, and BANA~\cite{oh2021background}, respectively. 
As shown in Figure~\ref{fig:pseudo}, there is a large gap, $10.78\%$ in the mIoU, between the performance of the SegNet trained with box annotation and the empirical upper-bound mIoU. 
The performance gap can be effectively reduced but still remains noticeable by using more accurate pseudo labels generated by GrabCut and BANA~\cite{oh2021background}.
This suggests noisy labels inevitable exist in pseudo labels and greatly hamper the performance of SegNet. 

We further incorporate re-weighting mechanism with manually-set ideal weights generated according to ground truth when training the three omni-supervised models. The resulting performance can be significantly improved (up to $8.71\%$ in the mIoU) and approaches to the upper-bound mIoU.
This accords with our design motivation that accurately identifying noisy labels pixel-wise and ignore them during training is the key to dealing with label noise and in turn improving segmentation performance.

\noindent\bd{Applicability to different frameworks.}
We verify university of our MetaSeg by applying it to different segmentation frameworks, including FCN~\cite{Zhao_2017_CVPR}, Deeplab V3 Plus (D3P)~\cite{deeplabv3plus2018}, and UperNet~\cite{Long_2015_CVPR}.
Figure~\ref{fig:method} shows our method consistently boosts the performance of FCN, Deeplab V3 Plus, and UperNet baseline by large margins, up to $3.35\%$ in the mIoU. It also outperforms their corresponding  fine-tuned counterparts by $1.65\%$, $2.56\%$ and $0.61\%$ in the mIoU. 
This suggests our method is applicable to different segmentation frameworks. In addition, the large performance improvement by MetaSeg is not caused by the introduction of additional training data ($\mathcal{D}_{\rm C}$), but is mainly attributed to the generated weights by CAM-Net that provide a better guidance to model optimization.

Similarly, we further verify the applicability of our MetaSeg to different backbones. We perform experiments based on FCN with VGG16~\cite{simonyan2014very}, ResNet18/50~\cite{he2016deep}, and HRNet18~\cite{wang2020deep}. 
Figure~\ref{fig:backbone} shows that compared to the fine-tuned baseline model, our method consistently boosts the performance of different backbones by up to $3\%$ in the mIoU.  
These results clear demonstrate that our MetaSeg is of high applicability can benefit a wide variety of segmentation frameworks and backbones.

\noindent\bd{Features used in CAM-Net.}
We incorporate three features in the CAM-Net: $\bm{F}_{u}$, $\bm E$ and $\bm P$. We add them to the CAM-Net one by one to investigate the effect of each individual feature.
Table~\ref{tab:feature} lists the ablation results. 
Incorporating $\bm{F}_{u}$ improves the mIoU by $1.15\%$ compared to the baseline model that does not introduce any features. 
This suggests $\bm{F}_{u}$ is useful in the sense of judging whether a region is overfitting to noisy labels according to the inherent inconsistency of its multi-level features. 
Incorporating $\bm E$ further improves the mIoU from $55.79\%$ to $55.96\%$. 
This improvement is due to the fact that CAM-Net can accurately identify noisy regions by comparing the difference between $\bm{F}_{l}$ and $\bm{F}_{u}$.
In addition, the performance improvement from $55.96\%$ to $56.29\%$ caused by introducing $\bm P$ well validates its effectiveness in strengthening features and thus highlighting inherent inconsistency between high-level and low-level features of noisy regions.

\begin{table}[t]
  \centering
  \renewcommand\arraystretch{1.2}
  \setlength{\tabcolsep}{14pt}
   \caption{Effect of the features used in CAM-Net.}
  \label{tab:feature}%
    \begin{tabular}{ccc|c}
        \toprule
     $\bm{F}_{u}$ & $\bm E$  & $\bm P$ &  mIoU (\%) \\
        \midrule
          &   &  & 54.64 \\
          \checkmark &   &  & 55.79 \\
          \checkmark &  \checkmark &  & 55.96 \\
           \checkmark &  \checkmark & \checkmark  & \bd{56.29} \\
        \bottomrule
    \end{tabular}
\end{table}

\noindent\bd{Effect of decoupled training strategy.} 
To verify the accelerated training effectiveness of the decoupled training strategy,  we train multiple models using different numbers of decoupled layers. 
We take a bottleneck block as a decoupled layer. Figure~\ref{fig:decouple} shows that as the number of decoupled layers increases, the mIoU steadily grows and the required training time decreases.
When using four decoupled layers ($S=4$), the mIoU is significantly improved by $3.51\% $ and the training speed is accelerated by $6.64$ times compared to the model without using decoupled training strategy ($S=0$).
The performance improvement and training acceleration are close to saturation at $S=4$. 
One possible explanation is that more fine-grained derivation of a small number of data in $\mathcal{D}_{\rm C}$ may lead to overfitting.

\noindent\bd{Impact of the scale of meta set.}
The meta set $\mathcal{D}_{\rm C}$ comprises a fixed number of images with accurate pixel-level masks. 
We investigate the impact of the number of images (scale) of the meta set by training multiple models using different proportions of images in $\mathcal{D}_{\rm C}$.
Figure~\ref{fig:proportions} shows the performance grows steadily as the proportions of the images used in $\mathcal{D}_{\rm C}$ increases. 
This indicates the more clean data used in $\mathcal{D}_{\rm C}$, the more accurate guidance given during model training and the better the performance achieved. 
Notably, by using only $20\%$ images in $\mathcal{D}_{\rm C}$, our MetaSeg achieves similar performance to the baseline model trained on $\mathcal{D}_{\rm B}$ and fine-tuned on the entire $\mathcal{D}_{\rm C}$.
The explanation is that our MetaSeg uses a small amount of clean data in $\mathcal{D}_{\rm C}$ to learn to maximize the usage of the large amount of noisy data in $\mathcal{D}_{\rm B}$, rather than simply learning to fit the data in $\mathcal{D}_{\rm C}$ as the baseline model does. Consequently, our MetaSeg has high data-efficiency for well-annotated training data.

\noindent\bd{Impact of the number of category prototypes.}
To investigate relationship of the effectiveness of embedding class  feature and the numbers of prototypes, we train multiple models using different numbers of prototypes.
Figure~\ref{fig:attr} shows that the mIoU first gradually improves as the $R$ increases from $2$ to $8$ then drops when more category prototypes are introduced.
This suggests more category information can be embedded into class embbeding feature as the the prototypes increasing, and then brings increased performance.
However, too many category prototypes easily lead to overfitting and thus cause performance degradation.

\section{Conclusion}
In this work, we rethink omni-supervised semantic segmentation as a problem of learning with noisy labels.
As a first step, we introduce meta learning and design a novel CAM-Net that learns to generate per-pixel weights to suppress noisy regions while highlighting clean ones. Then pixel-wise re-weighting mechanism is adopted to guide the training of SegNet.
To solve the problem of slow training involved for meta learning, we introduce a new decoupled training strategy to accelerate the optimization of CAM-Net.
Extensive experiments demonstrate that our method achieves state-of-the-art performance on various segmentation tasks. We hope this work will pave a new way for omni-supervised semantic segmentation.

\bibliographystyle{IEEEtran}
\bibliography{MetaSeg}

\end{document}